\documentclass[conference]{IEEEtran}

\IEEEoverridecommandlockouts
\usepackage{cite}
\usepackage{amsmath,amssymb,amsfonts}
\usepackage{algorithmic}
\usepackage{graphicx}
\usepackage{textcomp}
\usepackage{xcolor}
\usepackage[tableposition = top]{caption}
\usepackage[T1]{fontenc}
\usepackage{hyperref}
\usepackage{float}
\usepackage{afterpage}
\usepackage{mwe}
\usepackage{subcaption}
\usepackage[export]{adjustbox}
\usepackage[T1]{fontenc}

\begin{document}
 \title{AA-TransUNet: Attention Augmented TransUNet For Nowcasting Tasks}



\author{\IEEEauthorblockN{Yimin Yang and Siamak Mehrkanoon*\thanks{*corresponding author.}}
\IEEEauthorblockA{\textit{Department of Data Science and Knowledge Engineering} \\
\textit{Maastricht University}\\
Maastricht, The Netherlands \\
yimin.yang@student.maastrichtuniversity.nl, siamak.mehrkanoon@maastrichtuniversity.nl}
}

\maketitle

\begin{abstract}

Data driven modeling based approaches have recently gained a lot of attention in many challenging meteorological applications including weather element forecasting. This paper introduces a novel data-driven predictive model based on TransUNet for precipitation nowcasting task. The TransUNet model which combines the Transformer and U-Net models has been previously successfully applied in medical segmentation tasks. Here, TransUNet is used as a core model and is further equipped with Convolutional Block Attention Modules (CBAM) and Depthwise-separable Convolution (DSC). The proposed Attention Augmented TransUNet (AA-TransUNet) model is evaluated on two distinct datasets: the Dutch precipitation map dataset and the French cloud cover dataset. The obtained results show that the proposed model outperforms other examined models on both tested datasets. Furthermore, the uncertainty analysis of the proposed AA-TransUNet is provided to give additional insights on its predictions.
\end{abstract}

\IEEEpeerreviewmaketitle

\section{Introduction}
Accurate prediction of weather conditions impacts human life and it has a direct effect on many sectors including for instance economic, agriculture, business, transport and logistics among others. In particular, precipitation nowcasting, i.e. short-term forecasts of up to 6 hours, is becoming an increasingly popular research topic due to its high impact in supporting the socioeconomic needs of the above-mentioned industrial sectors \cite{hall1999precipitation}. In addition, precipitation nowcasting plays an important role in water-related risk management. Often warning systems also rely on nowcasts in order to generate warnings \cite{de2013rainfall}.

Traditional nowcasting methods are based on Numerical weather prediction (NWP) \cite{bauer2015quiet}. However, their spatial and temporal resolution is often too low for precipitation nowcasting. In addition, NWP based models demand a substantial amount of computing power as they generate predictions by simulating the underlying physics of the atmosphere and ocean \cite{mehrkanoon2019deep,trebing2021smaat}. Therefore, recent years have witnessed the development of new data driven models to overcome the shortcomings of the NWP based models. They often do not rely on the underlying physics of the atmosphere and ocean but instead concentrate solely on learning from the historical data \cite{fernandez2021broad}. In particular, models with deep neural network architecture have shown to outperform traditional data driven models in several application domains \cite{yegnanarayana2009artificial,goodfellow2016deep,scher2018toward, mehrkanoon2019deep,trebing2020wind,stanczyk2021deep,
aykas2021multistream,abdellaoui2021symbolic,fernandez2021broad, bilgin2021tent, fernandez2020deep}.

Convolutional Neural Networks (CNN) based models such as AlexNet \cite{krizhevsky2012imagenet}, ResNet \cite{he2016deep}, Convolutional Autoencoders (CAEs) \cite{berthomier2020cloud} and their variants are among the most used neural networks architectures in computer vision tasks. Following the idea of CAEs, the authors in \cite{ronneberger2015u} introduced the well-known UNet model, which uses the classic encoder (down-sampling) and decoder (up-sampling) with skip connection structure \cite{ronneberger2015u}. UNet was initially used for medical image segmentation, but it gradually began to shine in many other computer vision tasks. Inspired by the success of UNet model, several other architectures have been developed in the literature which continues to adhere to the core concept of UNet, albeit incorporating new modules or other techniques. 

Another impactful model is the Transformer architecture \cite{vaswani2017attention} which has revolutionized the field of Natural Language Processing (NLP).
This sequence to sequence architecture which is equipped with attention mechanism is designed to handle sequential input data. Compared to traditional models used for sequence-to-sequence learning, Transformer has produced better results \cite{vaswani2017attention}. With the transformer model dominating the domain of NLP, many researchers began to investigate its applicability in computer vision tasks. For instance, Google researchers released the Vision Transformer model \cite{dosovitskiy2020image} which divides the image into small blocks of fixed size, linearly embeds each small block, adds position embedding and feeds the resulting vector sequence into a standard Transformer encoder. 

Although the Transformer based model has great potential in the computer vision field, it also faces some significant challenges such as the high demand for data \cite{nogueira2021investigating}. Likewise, the UNet structure that has consistently held a dominant position in the field of image segmentation also poses some limitations such as the inability to capture the long-term dependence \cite{chen2021transunet}.
Consequently, combining the transformer and the UNet model has become a new research direction. Based on this, the authors in \cite{chen2021transunet} proposed the TransUNet, which is an image segmentation model that fully utilizes the advantages of UNet and Transformer. The TransUNet model has been shown to perform well in medical image segmentation, but it has not been widely used in other fields. 

In this paper, we propose a new model called Attention Augmented TransUNet (AA-TransUNet). Here, the TransUNet is used as core model and it is further equipped with Convolutional Block Attention Module(CBAM) \cite{woo2018cbam} and Depthwise-separable Convolutions (DSC) \cite{chollet2017xception}. Thanks to the incorporation of these new elements in the core TransUNet model, the number of trainable parameters is significantly reduced without degrading the accuracy of the model. We seek the application of the proposed AA-TransUNet on precipitation nowcasting tasks. In particular, here we use the same datasets as those used in \cite{trebing2021smaat}, i.e precipitation map dataset and cloud cover dataset. As in  \cite{trebing2021smaat}, in the case of the precipitation map dataset, the model input consists of 60 minutes of precipitation maps collected from the cartographic radar and the model output is another sequence of precipitation maps predicted for the next 30 minutes. We also run related experiments on the cloud cover dataset to demonstrate the applicability of the proposed model. 

This paper is organized as follows. A brief overview of the related research works is given in Section~\ref{sect:rw}. Section~\ref{sect:pm} introduces the proposed AA-TransUNet model. The experimental settings and description of the used datasets are given in Section~\ref{sect:e}. The obtained results are discussed in Section~\ref{sect:rd} and the conclusion is drawn in Section~\ref{sect:c}.

\section{Related works}
\label{sect:rw}
The data driven based models have gained a lot of attentions for weather elements forecasting such as temperature, wind speed, and precipitation among others. Weather elements forecasting can be cast into sequence prediction problem and hence it is closely linked to Recurrent Neural Networks(RNN) models such as Long Short-term Memory (LSTM) \cite{hochreiter1997long}. In addition due to the presence of spatial structure of the weather data, it is also linked to convolutional Neural Networks (CNN). Therefore, recent years have witnessed the emergence of powerful deep learning architectures based on LSTM, CNN, and UNet for addressing challenging tasks in weather elements forecasting. 

For instance, the authors in \cite{xingjian2015convolutional} proposed the Convolutional LSTM (ConvLSTM) model that better captures the underlying spatiotemporal correlations of the data and consistently outperforms normal LSTM. Continuing in the ConvLSTM research direction, the authors in \cite{kumar2020convcast} proposed Convcast, an embedded convolutional LSTM-based architecture. The author in \cite{mehrkanoon2019deep} introduced different CNN architectures including 1-D, 2-D and 3-D convolutions to accurately predict wind speed in the next 6 to 12 hours. The UNet architecture is used for precipitation nowcasting in \cite{agrawal2019machine}. The SmaAt-UNet model, an extension of UNet model, which greatly reduces the UNet parameters without compromising its performance is introduced in \cite{trebing2021smaat}. The authors in \cite{fernandez2021broad}, introduced Broad-UNet by equipping the UNet model with asymmetric parallel convolutions as well as Atrous Spatial Pyramid Pooling (ASPP) module. 

Despite its many advantages, UNet can still result in limited abilities when modeling long-range dependencies due to the intrinsic locality of the convolution based operations. On the other hand, Transformer model, which is initially developed for sequence-to-sequence prediction and uses the global self-attention mechanisms \cite{vaswani2017attention}, has also limitations in localization abilities due to insufficient low-level details \cite{chen2021transunet}. Subsequently, the authors in \cite{chen2021transunet} proposed TransUNet model that has the best of two UNet and Transformer models and they showed its successful application in medical image segmentation tasks.

In this paper, we use TransUNet as the core model and extend it by incorporating Convolutional Block Attention Modules (CBAM) and employing the Depthwise-separable Convolutions (DSC) in the core model. 
CBAM is a lightweight and general attention module for feed-forward Convolutional Neural Networks. Given an intermediate feature map, CBAM infers attention maps in two dimensions (channel attention module and spatial attention module) sequentially \cite{woo2018cbam}. Depthwise (DW) convolution and Pointwise (PW) convolution are collectively called Depthwise-separable Convolution. This structure is similar to regular convolution operations and can be used to extract features,  although its parameter sizes are substantially smaller than standard convolution operations \cite{chollet2017xception}.

\bigskip
\section{Method}
\label{sect:pm}
This section introduces the proposed AA-TransUNet model which uses the TransUNet \cite{chen2021transunet} as the core model and extends it to reduce its parameters and improve its forecasting performance. We then investigate the application of the proposed model in precipitation nowcasting tasks.
\begin{figure*}
\centering
\includegraphics[scale=0.67]{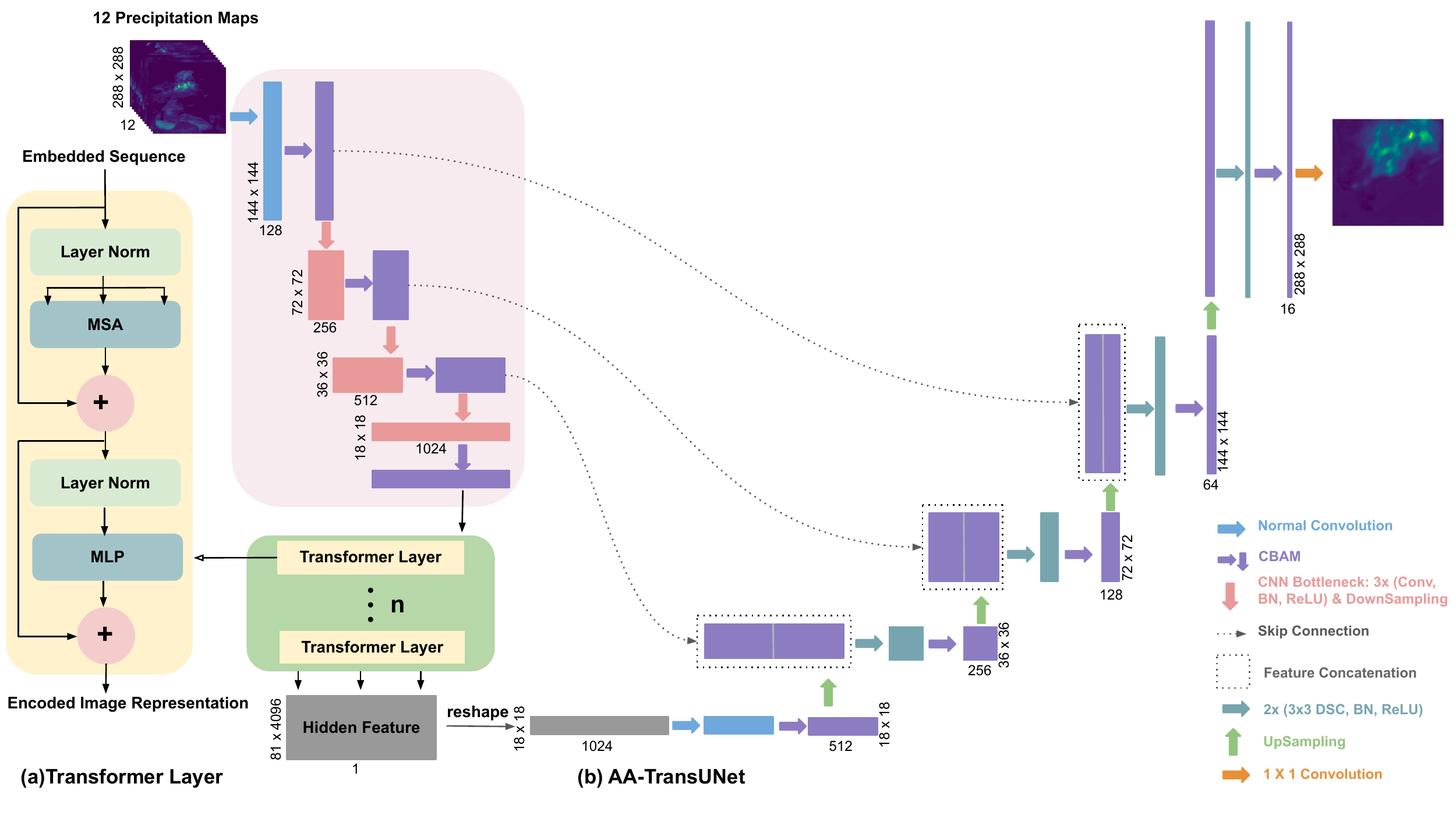}
\caption{The architecture of AA-TransUNet. (a) depicts the deconstruction schematic of Transformer Layer. (b) depicts the proposed model. Each bar in the figure represents a multi-channel feature map. The numbers below each bar display the number of channels and the numbers next to the bar represent the resolution. The input is the ground truth of an exemplary precipitation map, whereas the output is the corresponding prediction map produced by AA-TransUNet.}
\label{fig:p}
\end{figure*}

\subsection{Proposed Model}

The present study builds upon and extends the work of \cite{chen2021transunet}, who introduced TransUNet. The original TransUNet is an encoder-decoder architecture where CNN-Transformer is used as the encoder, while the original UNet decoder is utilized as the decoder. This model has been successfully applied in medical image segmentation tasks. 
Here, we propose a new encoder-decoder architecture called AA-TransUNet which equips the classical TransUNet with two key elements: Convolutional Block Attention Modules (CBAM) and Depthwise-separable Convolutions (DSC). The model is shown in Fig. \ref{fig:p} and is used for precipitation forecasting tasks. We integrate CBAM, which consists of a Channel Attention Module (CAM) and a Spatial Attention Module (SAM), to both encoder and decoder path of the TransUNet model. In particular, in the encoder path, CBAM is incorporated into the CNN part of the hybrid CNN-Transformer layer, and in the decoder path it is placed after all convolution layers (see Fig. \ref{fig:p}). This allows the model to perform both channel-wise and spatial-wise attention and therefore can better explore the inter-channel as well as inter-spatial relationship of the features.
In addition, the original convolutions of the decoder of the TransUNet is replaced by DSC aiming to reduce their parameter size.
In Table \ref{tab:para}, we compare the number of parameters of the decoder of TransUNet and AA-TransUNet models. Despite the added CBAM layers, the number of parameters of AA-TransUNet decoder are still 68.94\% lower than that of TransUNet. 
It should be noted that here we only compare the number of parameters of the decoder. As CBAM, which is a lightweight and general module \cite{woo2018cbam}, is only added to the encoder, therefore the number of parameters of the encoder path of the AA-TransUNet is relatively the same as that of TransUNet. 

\begin{table}[h]
\setlength{\tabcolsep}{1.6mm}
\caption{Comparison of the number parameters of the decoder of the two models.}
\begin{tabular}{cccc}
\hline
\textbf{Model}& \textbf{CBAM layers} & \textbf{Conv layers} & \textbf{Total Parameters}\\
\hline
TransUNet & 0 & $\sim$2.93M &$\sim$2.93M\\
AA-TransUNet & $\sim$0.04M & $\sim$0.87M & $\sim$0.91M\\
\hline
\end{tabular} 
\centering
\label{tab:para}
\end{table}

Despite \cite{trebing2021smaat}, here the input to the encoders is the image feature with the attention mechanism applied. In addition, as oppose to \cite{trebing2021smaat}, CBAM is used in each decoder level (see purple arrows in the decode path of Fig. \ref{fig:p}) to identify important features across channels and spatial dimensions of an image \cite{trebing2021smaat, woo2018cbam}.

Similar to TransUNet \cite{chen2021transunet}, the encoder of the AA-TransUNet model is a Transformer based model, which is typically used to process sequence data. Therefore, when using it as an encoder, the image sequentialization step needs to be taken first to covert 3D images to 2D patches \cite{chen2021transunet}. The obtained patches are then fed into the Transformer layer, which consists of Multihead Self-Attention (MSA) layers \cite{vaswani2017attention} and Multi-Layer Perceptron (MLP) blocks (each MSA and MLP pass through the layer normalization gate). At the same time, following \cite{chen2021transunet}, instead of using a pure Transformer as an encoder, we employ CNN-Transformer to make use of the intermediate high-resolution CNN feature maps in the decoding path and improve the model performance.

An architecture similar to UNet decoder is used for the proposed AA-TransUNet decoder. As shown in Fig. \ref{fig:p}, four operations are used in the AA-TransUNet decoder: upsampling (green arrows), 
feature concatenation (dotted rectangles),
double convolution (cyan arrows) 
and finally the CBAM (purple arrows). The last layer of the model is a 1 x 1 convolution (orange arrow), which produces a single feature map that represents the prediction results. 

\subsection{Training}
We have compared our proposed AA-TransUNet model with four other models, i.e. persistence method, UNet \cite{ronneberger2015u}, SmaAt-UNet \cite{trebing2021smaat} and TransUNet \cite{chen2021transunet} models on two datasets: precipitation maps as well as cloud cover dataset. The persistence method, a commonly used approach in nowcasting tasks, refers to the model that uses the last input image of a sequence as the prediction image. The other four examined UNet based models are trained under the same settings. The maximum number of epochs for precipitation maps dataset and cloud cover dataset is set to 200 and 100 respectively.
Following the lines of \cite{trebing2021smaat}, an early stopping criterion method is adopted to improve training efficiency and avoid overfitting. In this case, the model will cease training if the validation set does not increase in the last 20 epochs. 
For initializing the learning rate a range of values, i.e 0.01, 0.005, 0.001, 0.0005 and 0.0001 have been tested and we empirically found that 0.001 was the optimal choice to be set as the initial learning rate. We employ a learning rate scheduler that reduced the learning rate to a tenth of the previous learning rate when the validation loss does not increase for four epochs. The Adam optimizer is used and the batch size is set to 6. The Mean Squared Error (MSE) is used as loss function. Different number of Transformer layers, i.e. 1, 3, 6, 12 and 18 layers, have been examined. However, in all the conducted experiments, the number of Transformer layer in both TransUNet and AA-TransUNet is set to 1 as it was empirically found to be the optimal choice. All the compared models are trained using Pytorch on NVIDIA Tesla P100 with 16GB of RAM. 
The Pytorch implementation of the models is available at GitHub \footnote{\url{https://github.com/YangYimin98/AA-TransUNet}}.

\subsection{Model Evaluation}
The results of five metrics including Mean Squared Error (MSE), Accuracy, Precision, Recall and F1-score are reported to evaluate the performance of the models. Similar to \cite{trebing2021smaat}, in order to calculate true positives (TP), false positives (FP), true negatives (TN) and false negatives (FN), each pixel of the model output and the target image are converted to a boolean mask by comparing with a fixed threshold. The threshold value is set to 0.5 in all test cases.

\section{Experiments}
\label{sect:e}
In this section, the details of the conducted experiments on these two datasets, i.e. precipitation maps and cloud cover datasets are given. We use the same datasets as those used in \cite{trebing2021smaat}.

\subsection{Precipitation maps nowcasting}
The data is from the Royal Netherlands Meteorological Institute (Koninklijk Nederlands Meteorologisch Instituut, KNMI) and it has previously been used in \cite{trebing2021smaat}. It covers the precipitation information in the Kingdom of the Netherlands and its neighboring countries from 2016 to 2019. The data was collected using two C-band Doppler weather radar stations located in De Bilt and Den Helder. The satellite collects precipitation data every five minutes and stores it in the image format. The total number of precipitation maps is 420,000. Each raw image is 756 x 700 in size and the value of each pixel represents the total amount of precipitation collected per square kilometer in the past five minutes and is stored in an integer format. The details of the dataset are given in \cite{trebing2021smaat} and the dataset can be found at GitHub \footnote{\url{https://github.com/HansBambel/SmaAt-UNet}}.

\begin{figure}[htbp]
\centering
\includegraphics[scale=0.7]{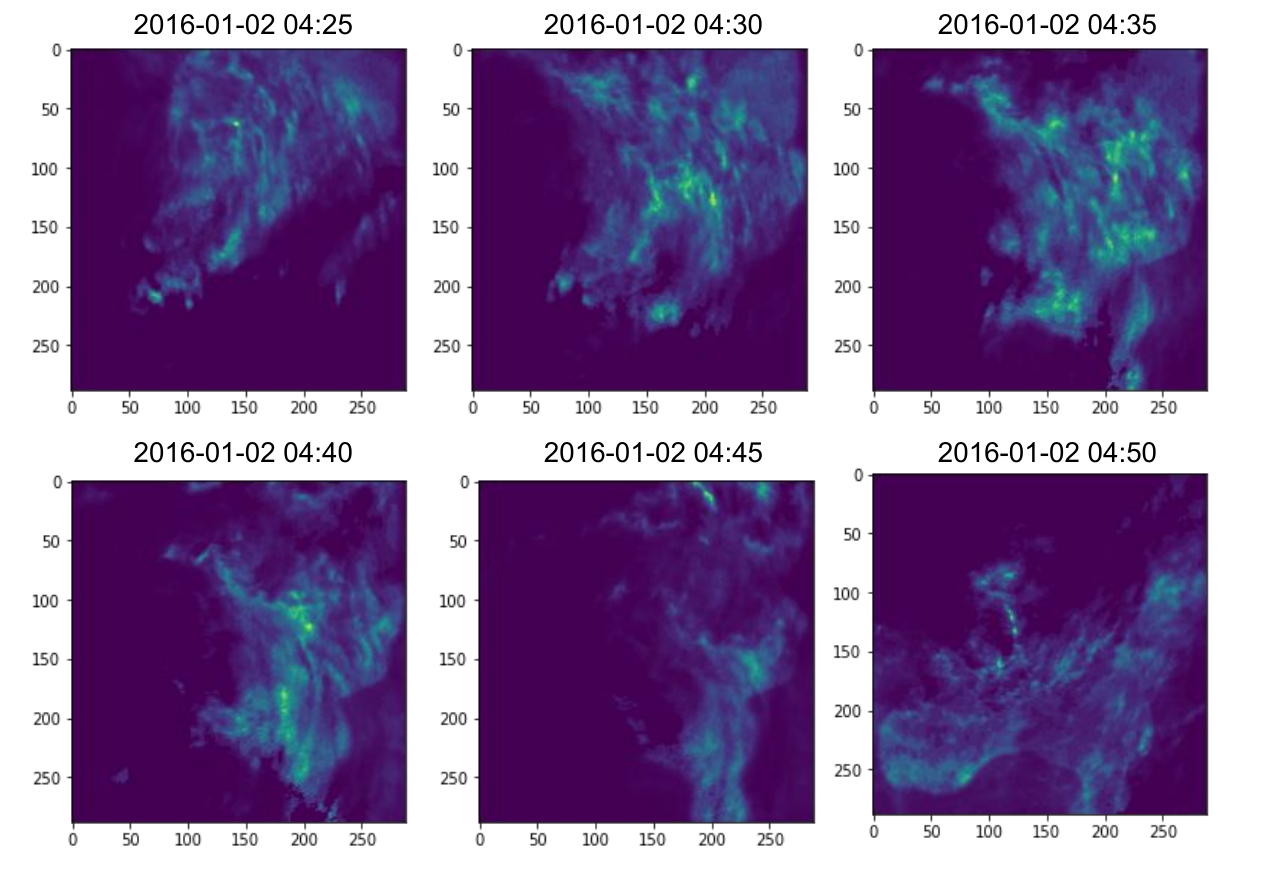}
\caption{Example of NL-50 dataset over 30 minutes.}
\label{fig:prec}
\end{figure}

The same pre-processing steps as those used in \cite{trebing2021smaat} are also applied here. All the images in the precipitation dataset are cropped to 288 x 288 so that the model can focus more on the original image containing the rain pixels. Moreover, there is a high imbalance between pixels with and without rain, with many images lacking raining pixels. Therefore, as in \cite{trebing2021smaat}, we created two datasets, i.e. NL-50 and NL-20, by filtering the original precipitation dataset and choosing only the images with at least 50\% and 20\% of pixels containing any amount of rain respectively. 
The information of the datasets before and after cropping can be found in Table \ref{tab:dataset inf}. Fig. \ref{fig:prec} shows an example of NL-50 dataset over 30 minutes. 

As a pre-processing step, we normalize both NL-50 and NL-20 datasets by dividing them by the highest value of the NL-50 training set. Each data sample is a sequence of 18 images. In order to make a fair comparison, similar to \cite{trebing2021smaat}, we use the first 60 minutes data (first 12 images) as the model input and the last 30 minutes data as the target image for comparison with the model predictions. Additionally, the dataset from 2016 to 2018 is used as training set and the dataset of 2019 is used as test set. 10\% of the training set is randomly selected as a validation set to hint when to stop training.

The training set of NL-50 is used to train the models. The validation set of NL-50 is used for model selection. Once all the models are trained, they are tested on both test set of NL-50 as well as NL-20 datasets. In this way, the trained models are tested on different conditions. In particular, by testing the models on NL-20 test set, the generalization ability of the models are examined.

\begin{table}[h]
\setlength{\tabcolsep}{1.59mm}
\caption{The size of three precipitation datasets. Only NL-50 is used for training the models. The trained models are tested on both NL-50 and NL-20 test sets.}
\begin{tabular}{ccccc}
\hline
\textbf{Dataset}&  \textbf{Rain} & \textbf{Training Set} & \textbf{Validation Set} & \textbf{Test Set}\\
\textbf{Index} & \textbf{Pixels} & \textbf{(2016-2018)} & \textbf{(\%10 Training)} & \textbf{(2019)}\\\hline
Original & No Extraction & 314940 & 31494 & 105003 \\
NL-20 & 20\% pixels & 31674 & 3168 & 11276 \\
NL-50& 50\% pixels & 5734 & 573  & 1557 \\\hline
\end{tabular} 
\centering
\label{tab:dataset inf}
\end{table}

\subsection{Cloud cover nowcasting}
The cloud cover data is introduced in \cite{berthomier2020cloud} and previously used in \cite{trebing2021smaat}.
The cloud cover dataset is classified into 16 classes according to the height and type of the cloud. This classification is computed with various visible and infrared channel images shot by Meteosat Second Generation (MSG), a geostationary satellite at longitude 0 degrees from 2017 and 2018 in France. The data collected by the satellite every 15 minutes contains images of 3712 x 3712 size. 

In the cloud cover image, each pixel might have 15 different values. The details of the cloud cover states are reported in Table \ref{tab:cloud cover}. The data is binarized depending on whether there is a cloud cover in the image. More precisely, as in \cite{trebing2021smaat}, values of 1 to 4 is converted to 0 (no cloud) and values of 5 to 15 is converted to 1 (with cloud). Subsequently, the images are cropped into the size of 256 x 256 based on the French boundaries \cite{fernandez2021broad}.
\begin{table}[h!]
\setlength{\tabcolsep}{1.8mm}
\caption{The details of 15 possible values of every pixel, and the values after being binarized.}
\begin{tabular}{cccc}
\hline
\textbf{Values} & \textbf{Cloud Cover States} & \textbf{Binarized} \\ \hline
1 & Cloud-free land & 0 \\ 
2 &  Cloud-free sea & 0 \\ 
3 & Snow over land & 0 \\ 
4 &  Sea ice & 0 \\ 
5 &Very low clouds & 1 \\ 
6 &  Low clouds & 1\\ 
7 &   Mid-level clouds & 1\\ 
8 & High opaque clouds & 1 \\ 
9 & Very high opaque clouds & 1 \\ 
10 & Fractional clouds & 1 \\ 
11 & High semitransparent thin clouds & 1 \\ 
12 & High semitransparent meanly thick clouds & 1 \\ 
13 & High semitransparent thick clouds & 1 \\ 
14 &   High semitransparent above low or medium clouds & 1\\ 
15 &  High semitransparent above snow/ice & 1\\ \hline
\end{tabular}
\label{tab:cloud cover}
\end{table}

\begin{figure}[htbp]
\centering
\includegraphics[scale=0.43]{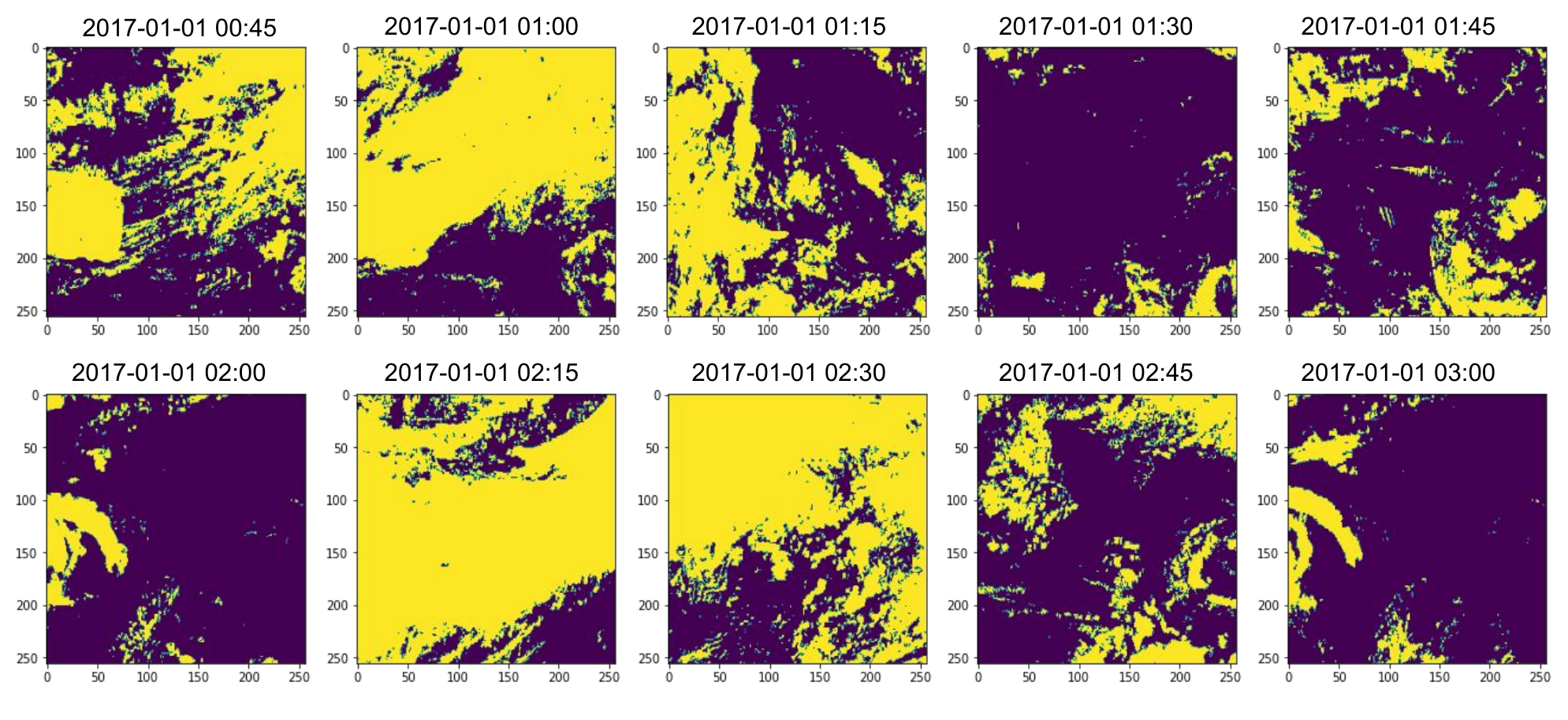}
\caption{An example of cloud cover data samples collected every 15 minutes. First 4 images are used as input and the remaining 6 images as output.}
\label{fig:cloud}
\end{figure}

Fig. \ref{fig:cloud}, shows a cloud cover data sample. Following the lines of \cite{trebing2021smaat}, we predict the cloud cover state for the last 90 minutes (last 6 images) based on the first 60 minutes (first 4 images). Similar to \cite{trebing2021smaat}, we use the data from 2017 and the first semester of 2018 as the training set and randomly select $10\%$ of them as the validation set. The remaining part of the dataset in 2018 is then used as a test set. Unlike the previous precipitation dataset processing method, the cloud cover dataset is not normalized since it only contains images with binary values of 0 and 1. 

\section{Results \& Discussion}
\label{sect:rd}
The obtained results of the examined models on the test set of three datasets, i.e. NL-50, NL-20 and cloud cover as well as their corresponding analysis are given in this section. 

\subsection{Evaluation of the precipitation maps nowcasting}
Table \ref{tab:nl50} shows the obtained results on the NL-50 test set. From Table \ref{tab:nl50}, it can be observed that all models outperform the persistent baseline model. It should be noted that as the weather data does not vary substantially in a short period of time (input is only 1 hour for both precipitation forecasting and cloud cover forecasting), the persistence model can already offer good prediction results. However, Table \ref{tab:nl50} shows that on the NL-50 test set, even the worst-performing UNet based model, i.e. TransUNet, has an MSE value less than half that of the persistence model.

The proposed AA-TransUNet model obtained the lowest MSE value compared to all the other examined models. The obtained MSE of TransUNet model is 0.0124 which is close to that of UNet and SmaAt-UNet. However, when Convolutional Block Attention Modules (CBAM) and Depthwise-separable Convolutions (DSC) are incorporated, the MSE score reduces to 0.0110, indicating that the proposed model improves over classical TransUNet by 11\%. In addition, note that after replacing the ordinary convolution by DSC in TransUNet decoder, the number of parameters of the decoder of TransUNet is greatly reduced. Moreover, besides the obtained MSE values, the proposed model also performs well on the other metrics.

\begin{table}[h]
\setlength{\tabcolsep}{1.5mm}
\caption{The obtained results of different metrics on NL-50 test set. The best results are presented in bold. Additionally, ↑ indicates that the higher the value of metrics, the better the performance, and ↓ indicates that the lower the value, the better the performance.}
\begin{tabular}{cccccc}
\hline
\textbf{Model} &  \textbf{MSE↓}  & \textbf{Accuracy↑}  & \textbf{Precision↑} & \textbf{Recall↑} & \textbf{F1↑}  \\ \hline
Persistence &0.0248 & 0.756 & 0.678 & 0.643 & 0.660\\
UNet & 0.0122 & 0.836 & 0.740 & \textbf{0.855} & \textbf{0.794} \\
SmaAt-UNet  & 0.0122& 0.829 & 0.730 & 0.850 & 0.786 \\
TransUNet  & 0.0124& 0.831 & 0.741 & 0.832 & 0.784 \\
AA-TransUNet & \textbf{0.0110} &\textbf{0.837} & \textbf{0.758} & 0.816  & 0.790 \\\hline
\end{tabular} 
\centering
\label{tab:nl50}
\end{table}

\begin{figure}[htbp]
\centering
\includegraphics[scale=0.4]{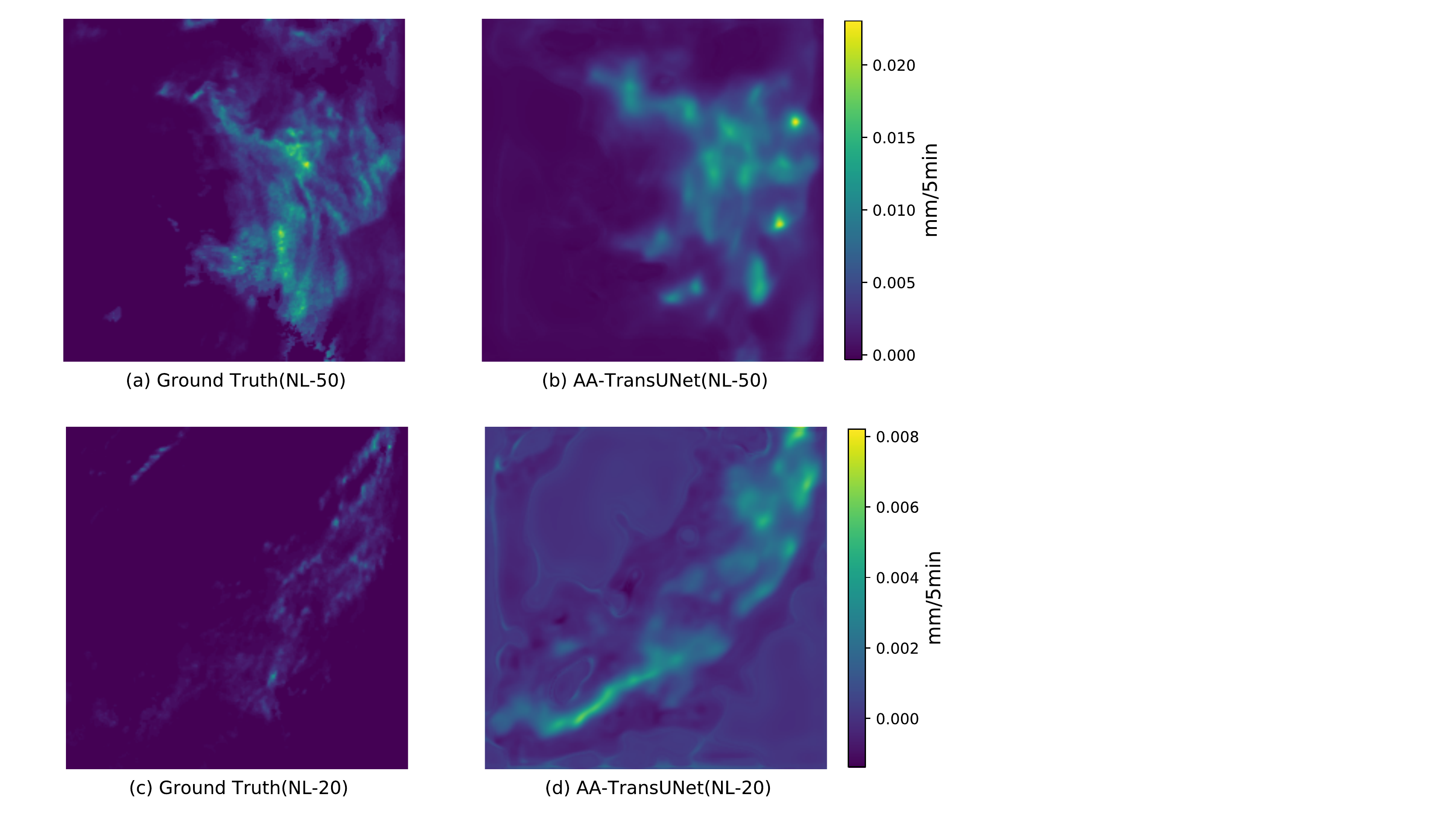}
\caption{Example of precipitation nowcasting of AA-TransUNet model. (a) and (b): The ground truth and the AA-TransUNet prediction for NL-50 dataset, respectively. (c) and (d): The ground truth and the AA-TransUNet prediction for NL-20 dataset, respectively.}
\label{fig:pred}
\end{figure}

An example of the ground truth sample images of NL-50 and NL-20 as well as the AA-TransUNet predictions are shown in Fig. \ref{fig:pred}. As it can be seen the sample image of NL-50 has significantly more rain pixels than that of NL-20. By comparing the ground truth and the model prediction, one can observe that our proposed model can accurately predict the concentrated locations of the rain pixels. At the same time, the prediction results of the model are a little blurry when compared to the ground truth. This is due to the fact that we use MSE as the loss function for training the model, and MSE loss tends to perform gradient descent in the direction of producing blurry images \cite{denton2018stochastic}.

We also test the models on the NL-20 test set. Table \ref{tab:dataset prec20} shows that our proposed AA-TransUNet model outperforms the other models on NL-20, despite the fact that the density of the rain pixels on NL-20 is much lower than that of NL-50.  

\begin{table}[h]
\setlength{\tabcolsep}{1.5mm}
\caption{The obtained results of different metrics on NL-20 test set. The best results are presented in bold. Additionally, ↑ indicates that the higher the value of metrics, the better the performance of the model, and ↓ indicates the lower the values, the better the performance.}
\begin{tabular}{cccccc}
\hline
\textbf{Model} &  \textbf{MSE↓}  & \textbf{Accuracy↑}  & \textbf{Precision↑} & \textbf{Recall↑} & \textbf{F1↑}  \\ \hline
Persistence &0.0227&	0.827&	0.559&	0.543&	0.551\\
UNet & 0.0111&	\textbf{0.880}&	0.666&	0.782&	\textbf{0.719} \\
SmaAt-UNet  &0.0111&	0.867&	0.626&	0.801&	0.703 \\
TransUNet & 0.0112&	0.872&	0.637&	\textbf{0.805}&0.705 \\
AA-TransUNet& \textbf{0.0106}&	\textbf{0.880}&	\textbf{0.671}&	0.760&	0.711\\\hline
\end{tabular} 
\centering
\label{tab:dataset prec20}
\end{table}

Table \ref{tab:dataset prec20} shows that the prediction results of all the models greatly exceed the results of the persistence model. Furthermore, the performance of AA-TransUNet, which obtains the lowest MSE value also achieves 5\% improvement compared to classical TransUNet. 

Additionally, We also give a comparison of the prediction map on NL-20 and the ground truth map. One can observe from Fig. \ref{fig:pred}(c, d) that despite the fact that the density of the ground truth rain pixels is very low, our model still predicts almost all of the contours of rain pixels.

\begin{figure}[htbp]
\centering
\includegraphics[scale=0.43]{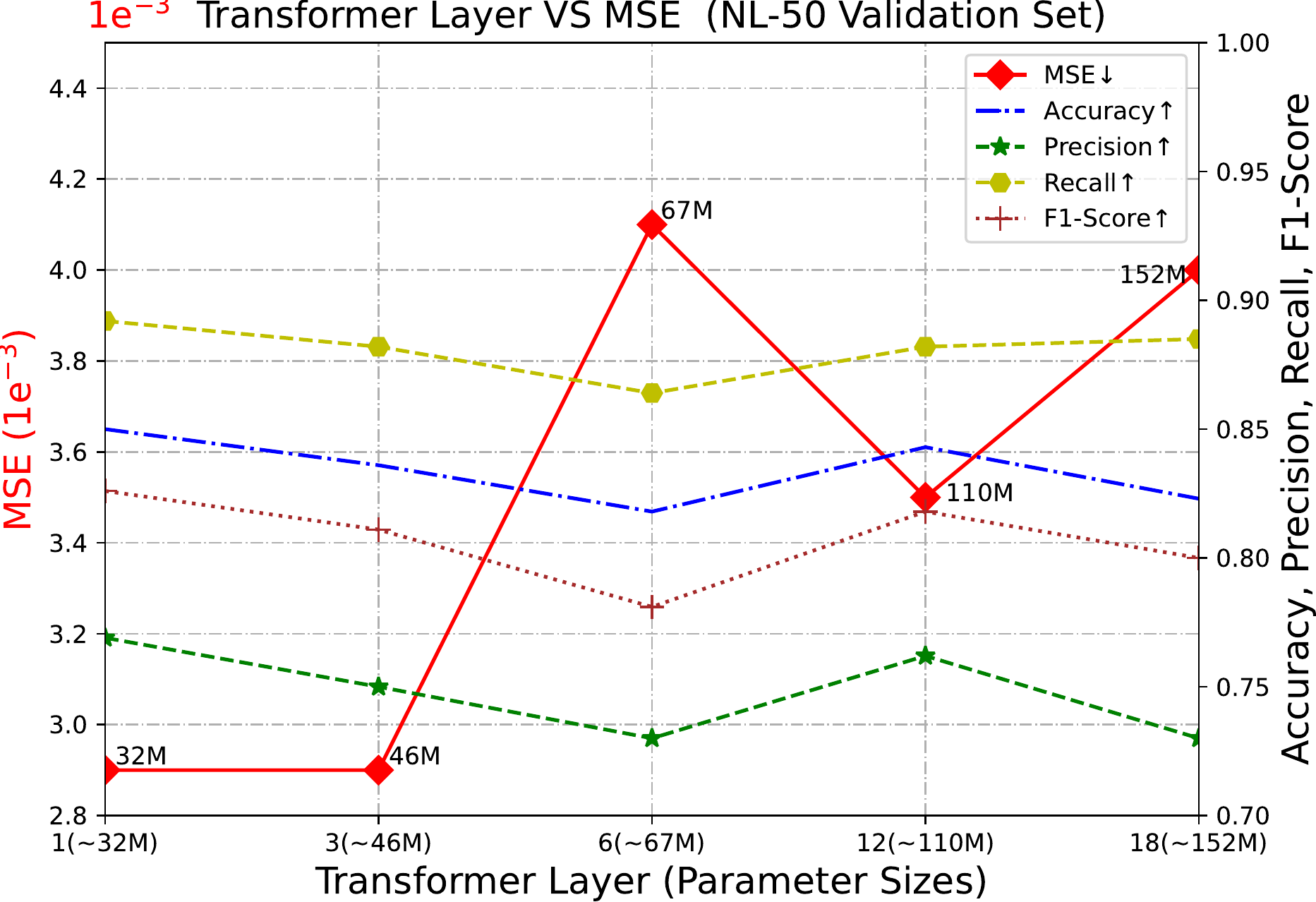}
\caption{The obtained MSE, Accuracy, Precision, Recall and F1-Score for NL-50 Validation Set using AA-TransUNet model with different Transformer layers. 
}
\label{fig:comp}
\end{figure}

Next, we analyze the influence of different number of Transformer layers on the performance of the AA-TransUNet model for the NL-50 test set. As it can be seen from Fig. \ref{fig:p}, each Transformer layer consists of Multihead Self-Attention (MSA) and Multi-Layer Perceptron (MLP) blocks. We have examined the cases where there are 1, 3, 6, 12 and 18 blocks (corresponding to 1, 3, 6, 12 and 18 Transformer layers). According to Fig. \ref{fig:comp}, the model with one Transformer layer (one MSA and one MLP block) performs the best on NL-50 validation set. This observation is in agreement with the findings of \cite{lee2021vision, neyshabur2020towards, zhou2021deepvit}, i.e. when the dataset size is small, using Transformer with fewer number of blocks (Transformer layers) can result in better performance. 

\subsection{Evaluation of the cloud cover nowcasting}
The above-mentioned models are trained and tested on the cloud cover dataset and the obtained results are tabulated in Table \ref{tab:cloud cover result}.
From Table \ref{tab:cloud cover result}, one can observe the same performance pattern as that of precipitation maps dataset. All the models have achieved results far exceeding the persistence model. At the same time, AA-TransUNet has the lowest MSE value and the highest value in accuracy and precision among all models. It is worth mentioning that the MSE value of all the models on the cloud cover data set is higher than the value previously obtained on the precipitation dataset since the cloud cover dataset only contains binary values. Therefore, the value of the actual label is sometimes very different from the predicted value, which consequently results in higher MSE values.

\begin{table}[h]
\setlength{\tabcolsep}{1.5mm}
\caption{The obtained results of different metrics for cloud cover test set. The best results is presented in bold. Additionally, ↑ indicates that the higher the value of metrics, the better the performance of the model, and ↓ indicates the lower the value, the better the performance.}
\begin{tabular}{cccccc}
\hline
\textbf{Model} &  \textbf{MSE↓}  & \textbf{Accuracy↑}  & \textbf{Precision↑} & \textbf{Recall↑} & \textbf{F1↑}  \\ \hline
Persistence &0.1491	&0.851	&0.849&	0.849&	0.849\\
UNet & 0.0785	&0.890&	0.895&	0.919&	\textbf{0.907} \\
SmaAt-UNet  &0.0794	&0.889&	0.892&	\textbf{0.921}&	0.906 \\
TransUNet & 0.0805&	0.887&	0.894&	0.914	&0.904 \\
AA-TransUNet& \textbf{0.0784}	&\textbf{0.891}&	\textbf{0.900}&0.913&0.906\\\hline
\end{tabular} 
\centering
\label{tab:cloud cover result}
\end{table}

\begin{figure}[htbp]
\centering
\includegraphics[scale=0.31]{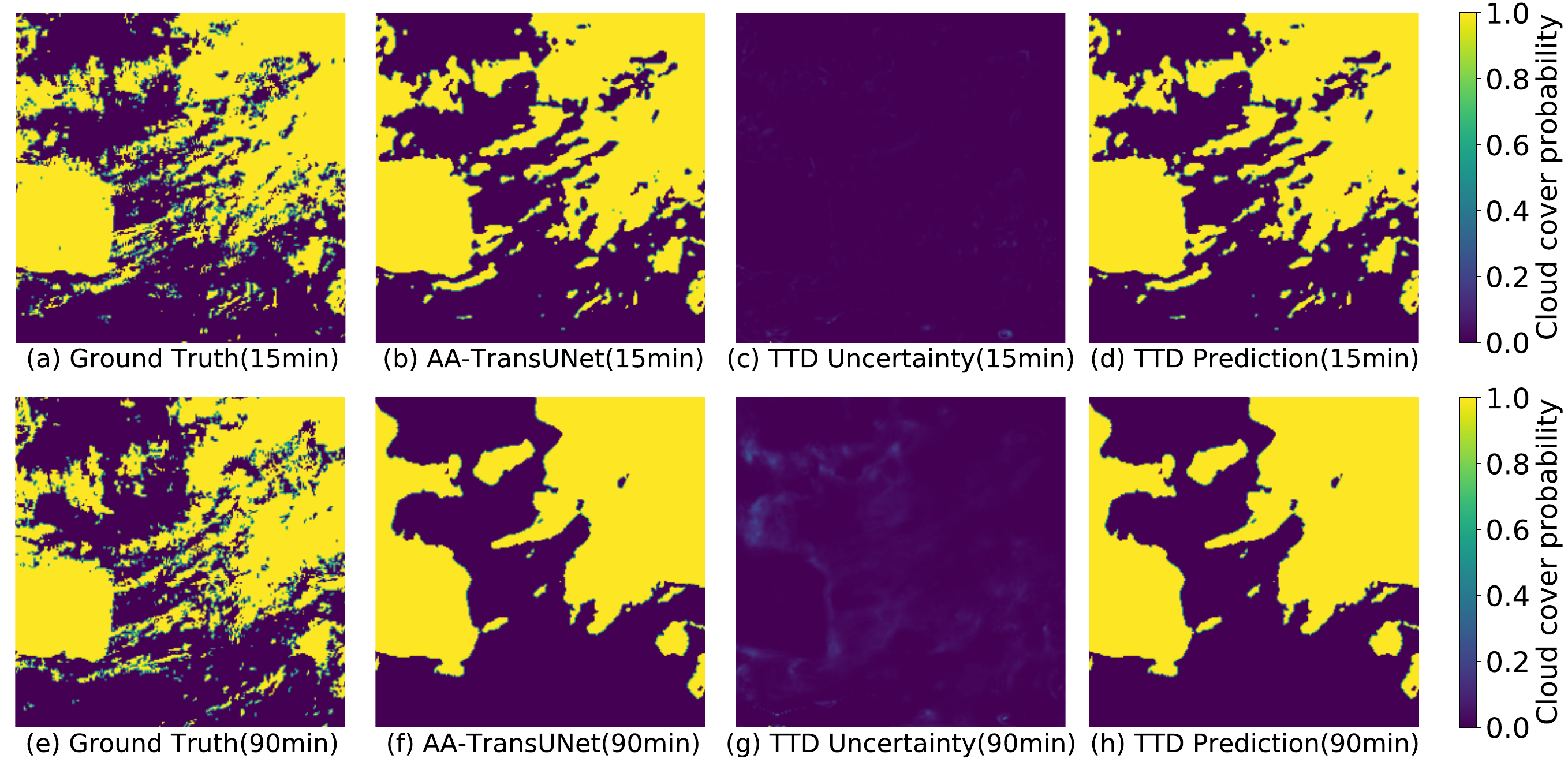}
\caption{Examples of cloud cover nowcasting using AA-TransUNet model. The first and second row correspond to 15mins and 90mins ahead predictions respectively. The four columns represent the Ground Truth, AA-TransUnet, TTD uncertainty map and TTD prediction respectively.}

\label{fig:pred cloud}
\end{figure}

An example of the AA-TransUNet prediction and the ground truth image of the cloud cover data is shown in Fig. \ref{fig:pred cloud}. The model has accurately identified all the appearances of the cloud cover on the 256 x 256 size image. 

\subsection{Uncertainty analysis}
\begin{figure}[htbp]
\centering
\includegraphics[scale=0.31]{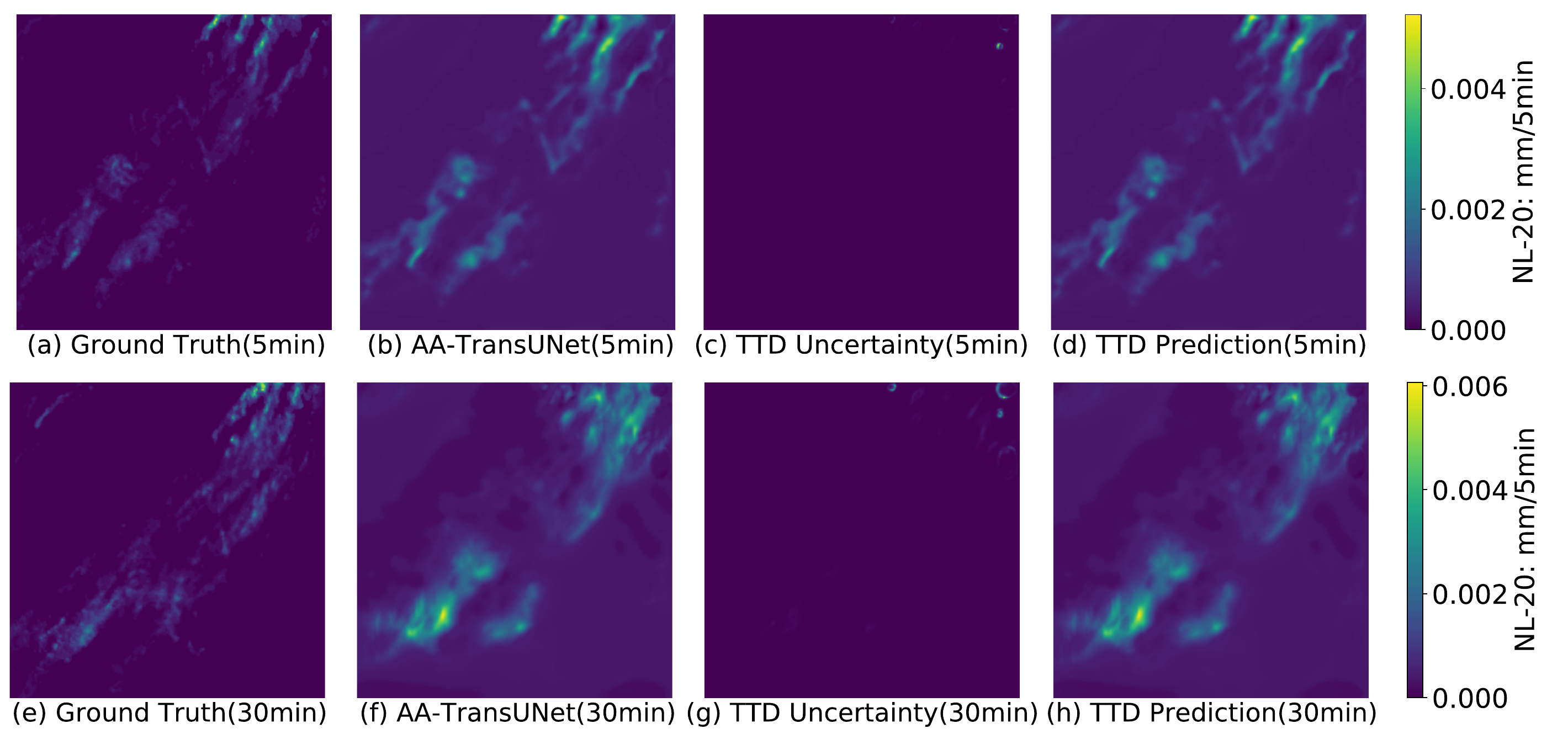}
\caption{Examples of precipitation nowcasting on NL-20 using AA-TransUNet model. The first and second row correspond to 5mins and 30mins ahead predictions respectively. The four columns represent the Ground Truth, AA-TransUnet, TTD uncertainty map and TTD prediction respectively.}
\label{fig:preds}

\end{figure}

\begin{figure}[htbp]
\centering
\includegraphics[scale=0.41]{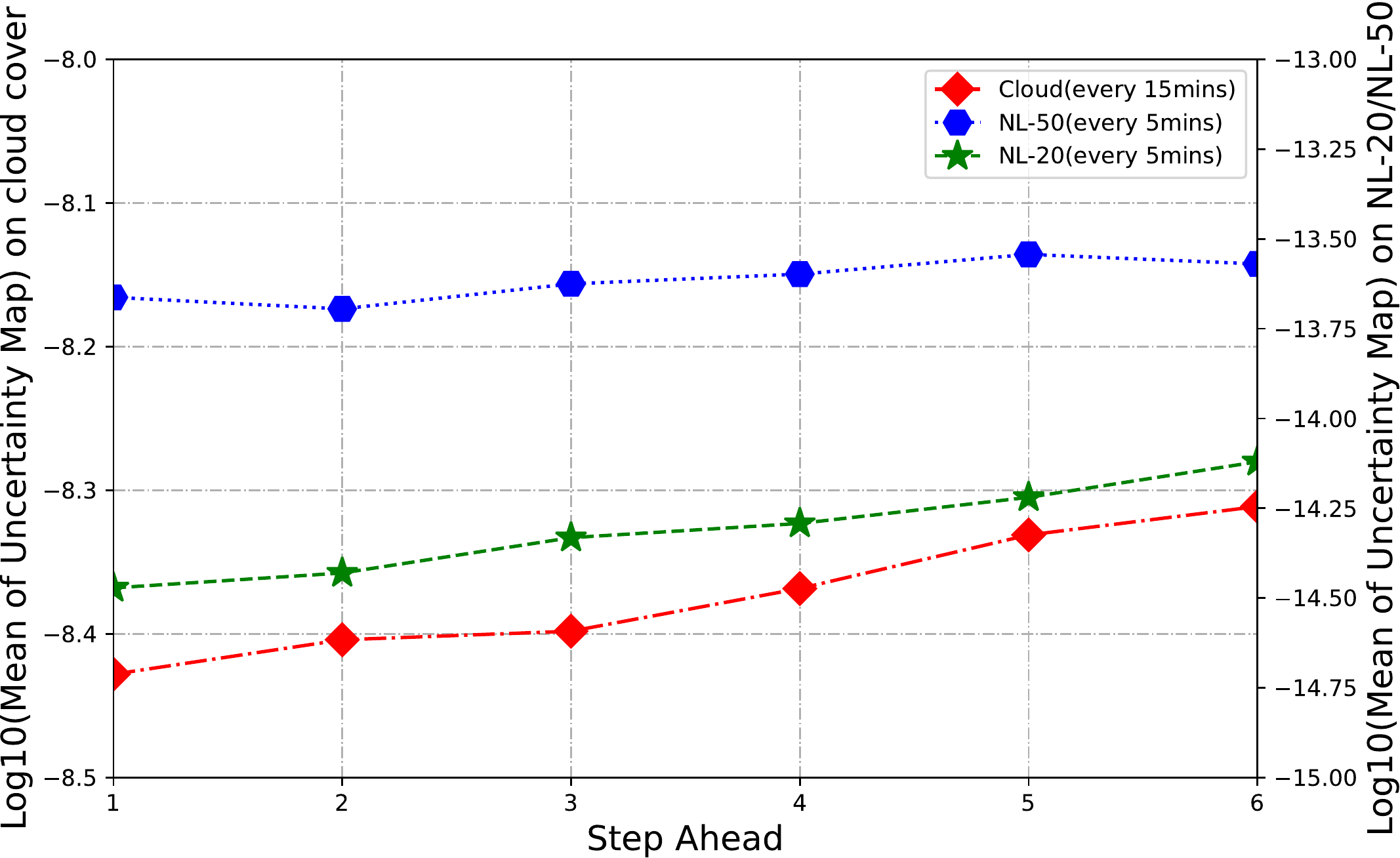}
\caption{The uncertainty quantification results of all the three datasets, i.e. NL-50, NL-20 and cloud cover dataset. The uncertainty value is obtained by taking the log10 of the mean of the uncertainty maps.}
\label{fig:unc}
\end{figure}

We quantify the uncertainty by means of Epistemic uncertainty which measures the uncertainty of the model predictions. In theory, Epistemic uncertainty can be obtained using Bayesian Neural Networks \cite{gal2016dropout}. However, as in \cite{gal2016dropout, natekar2020demystifying}, here we use test time dropout (TTD) to approximate the Bayesian inference.
In particular, at test time, a posterior distribution is generated by running the model for $k$ (in our case we set $k$=20) epochs for each target precipitation map. The uncertainty map of the model is then measured by taking the variance of the obtained $k$ precipitation maps (posterior sampled distribution) \cite{gal2016dropout}. In addition, we also take the mean of the posterior sampled distribution that can be considered as model prediction and is referred to as TTD prediction. The obtained TTD uncertainty map and TTD prediction for could cover as well as NL-20 datasets are shown in Fig. \ref{fig:pred cloud} and Fig. \ref{fig:preds}. Furthermore, Fig. \ref{fig:unc} shows that as the number of step ahead prediction increases, the uncertainty of the proposed AA-TransUNet model also increases.


\section{Conclusion}
\label{sect:c}
In this paper a novel Attention Augmented TransUNet model (AA-TransUNet) is proposed for precipitation nowcasting tasks. We show that incorporating the Convolutional Block Attention Modules (CBAM) and Depthwise-separable Convolution(DSC) into the classical TransUNet gives rise to better performance while significantly reducing the number of parameters in the decoder part of the model. The applicability of the proposed model is shown on two nowcasting tasks, i.e. precipitation as well as cloud cover nowcasting. The experimental results demonstrate that the AA-TransUNet model outperforms the original TransUNet model and other examined models. The implementation of our proposed model including the trained models is available at Github: \href{https://github.com/YangYimin98/AA-TransUNet}{https://github.com/YangYimin98/AA-TransUNet}.

\bibliographystyle{IEEEtran}
\bibliography{main}
\end{document}